\documentclass{article} 
\usepackage{nips15submit_e,times}
\usepackage{hyperref}
\usepackage{url}
\usepackage{authblk}
\usepackage{amsmath,graphicx,booktabs}
\usepackage{epstopdf,amssymb}
\usepackage{multirow}
\usepackage{fixltx2e}
\usepackage{subfigure}
\usepackage{float}
\usepackage{titlesec}
\usepackage{bm}

\title{\textbf{The Method of Multimodal MRI Brain Image Segmentation Based on Differential Geometric Features}} 
\author{Yongpei Zhu$^{1}$, Zicong Zhou$^{2}$, Guojun Liao$^{2}$, Qianxi Yang$^{1}$, Kehong Yuan$^{1*}$\\
	$^{1}$Graduate School at Shenzhen, Tsinghua University, Shenzhen 518055, China.\\
	$^{2}$The University of Texas at Arlington, Arlington 76019, USA.\\
	*Corresponding author: Kehong Yuan (e-mail: yuankh@sz.tsinghua.edu.cn)\\
	\texttt{zhuyp17@mails.tsinghua.edu.cn}\\
}

\nipsfinalcopy 

\begin{document}

\maketitle

\begin{abstract}
Accurate segmentation of brain tissue in magnetic resonance images (MRI) is a diffcult task due to different types of brain abnormalities. Using information and features from multimodal MRI including T1, T1-weighted inversion recovery (T1-IR) and T2-FLAIR and differential geometric features including the Jacobian determinant(JD) and the curl vector(CV) derived from T1 modality can result in a more accurate analysis of brain images. In this paper, we use the differential geometric information including JD and CV as image characteristics to measure the differences between different MRI images, which represent local size changes and local rotations of the brain image, and we can use them as one CNN channel with other three modalities (T1-weighted, T1-IR and T2-FLAIR) to get more accurate results of brain segmentation. We test this method on two datasets including IBSR dataset and MRBrainS datasets based on the deep voxelwise residual network, namely VoxResNet, and obtain excellent improvement over single modality or three modalities and increases average DSC(Cerebrospinal Fluid (CSF), Gray Matter (GM) and White Matter (WM)) by about 1.5\% on the well-known MRBrainS18 dataset and about 2.5\% on the IBSR dataset. Moreover, we discuss that one modality combined with its JD or CV information can replace the segmentation effect of three modalities, which can provide medical conveniences for doctor to diagnose because only to extract T1-modality MRI image of patients. Finally, we also compare the segmentation performance of our method in two networks, VoxResNet and U-Net network. The results show VoxResNet has a better performance than U-Net network with our method in brain MRI segmentation. We believe the proposed method can advance the performance in brain segmentation and clinical diagnosis.\\

\end{abstract}

\section{Introduction}
Magnetic resonance imaging (MRI) is usually the preferred method of structural brain analysis, as it provides high-contrast and high-spatial resolution images of soft tissue with no known health risks, which is the most popular choice to analyze the brain and we will focus on MRI in this work. Quantitative analysis of brain MR images is routine for many neurological diseases and conditions. Segmentation, i.e., labeling of pixels in 2D (voxels in 3D), is a critical component of quantitative analysis. Brain tissue segmentation generally refers to the separation of the brain into three functional components, namely, cerebrospinal fluid (CSF), grey matter (GM) and white matter (WM). There is a need for automated segmentation methods to provide accuracy close to manual segmentation with a high consistency.\\
 	
Deep learning techniques are gaining popularity in many areas of medical image analysis \cite{lin2016Neural}, different from traditional machine learning algorithm, it is a new and popular machine learning technique, which can extract complex feature levels from images. Some of the known deep learning algorithms are stacked auto-encoders, deep Boltzmann machines, deep neural networks, and convolutional neural networks (CNNs). CNNs are the most commonly applied to image segmentation and classification.\\	

At present, there are three main CNN architecture styles for brain MRI image segmentation \cite{Akkus2017Deep}: (1) Patch-Wise CNN Architecture. This is an easy way to train the CNN segmentation algorithm. Fixed size patches around each pixel were extracted from the given image, and then the training model was trained on these patches with pixel labels in the patch center, such as normal brain and tumor. The disadvantage of this method is that it is computationally intensive and difficult to train. (2) Semantic-Wise CNN Architecture \cite{Long2015Fully}\cite{Ronneberger2015U}. This architecture predicts each pixel of the entire input image, and the network only needs one forward inference. This structure includes the encoder part that extracts features and the decoder part that combines lower features from the encoder part to form abstract features. The input image is mapped to the segmentation labels in a way that minimizes a loss function. (3) Cascaded CNN Architecture \cite{Dou2016Automatic}. This type of architecture combines two CNN architectures. The first CNN is used for the preliminary prediction of the training model, and the second CNN is used to further adjust the prediction of the first network.\\	

Nowadays, the application of differential geometry in deep learning is more and more extensive, especially in the field of medical image. Based on the manifold characteristics of differential geometry, the success of deep learning is attributed to the inherent laws of the data itself. High-dimensional data are distributed near low-dimensional manifolds, which have a specific probability distribution, and it is also attributed to the strong ability of deep learning network to approximate nonlinear mapping. Deep learning technology can extract manifold structure from a kind of data and express the global prior knowledge with manifold, specifically, encoding and decoding mapping, which is implied in the weight of neurons. In the field of medical image analysis, doctors can determine whether the organs are abnormal by precisely comparing the geometry of the organs. By analyzing the geometric features of the tumor, we can judge the benign and malignant nature of the tumor. It can be attributed to the registration and analysis of medical images. Also the deep learning method based on differential geometry plays an important role in medical image registration.

Based on calculus of variation and optimization, we proposed a new
variational method with prescribed Jacobian determinant and curl vector to construct diffeomorphisms of MRI brain images. Since the Jacobian determinant has
a direct physical meaning in grid generation, i.e. the grid cell size changes, and the curl-vector represents the grid cell rotations, the deformation
method was applied successfully to grid generation and adaptation problems.

\section{Related Work}
Accurate automated segmentation of brain structures such as white matter (WM), gray matter (GM), and cerebrospinal fluid (CSF) in MRI is important for studying early brain development in infants and precise assessment of the brain tissue. In recent years, CNNs have been used in the segmentation of brain tissue, avoiding a clear definition of spatial and strength characteristics and providing better performance than the classical approach, which we will describe below.\\

Zhang et al. \cite{ZHANG2015214} proposed a 2D patch-wise CNN method to segment gray matter, white matter and cerebrospinal fluid from multimodal MR images of infants, and outperforms the traditional methods and machine learning algorithms; Nie et al.\cite{Nie2015FULLY} proposed a semantic-wise full convolution networks method and obtained improved results than Zhang's method. Their overall DSC were 85.5\%(CSF),87.3\%(GM), and 88.7\%(WM) vs. 83.5\%(CSF),85.2\%(GM), and 86.4\%(WM) by\cite{ZHANG2015214}; Moeskops et al. \cite{Moeskops2016Automatic} proposed a multi-scale (25$^{2}$,51$^{2}$,75$^{2}$pixels) patch-wise CNN method to segment brain images of infants and young adults with overall DSC=73.53\% vs. 72.5\% by \cite{Br2015Deep} in MICCAI challenge; De Brebisson et al.\cite{Br2015Deep} presented a 2D and 3D patch-wise CNN approach to segment human brain and they achieved competitive results(DSC=72.5\%$\mp$16.3\%) in MICCAI 2012 challenge. Bao et al. \cite{Bao2015Multi} also proposed a multi-scale patch-wise CNN method together with dynamic random walker with decay region of interest to obtain smooth segmentation of subcortical structures in IBSR (developed by the Centre for Morphometric Analysis at Massachusetts General Hospital-available at \footnote{\url{https://www.nitrc.org/projects/ibsr}} to download) and LPBA40 datasets; Chen et al. \cite{Chen2016VoxResNet} proposed deep voxelwise residual networks for volumetric brain segmentation, which borrows the spirit of deep residual learning in 2D image recognition tasks, and is extended into a 3D variant for handling volumetric data. Olaf Ronneberger et al.\cite{Ronneberger2015U} presents an architecture named U-Net which consists of a contracting path to capture context and a symmetric expanding path that enables precise localization, and it outperforms the prior best method(a sliding-window convolutional network) on the ISBI challenge for segmentation of neuronal structures.\\

In this paper, we use the differential geometric information including JD and CV, which represent the change rate of the area or volume of the brain image, and we can use them as one CNN channel with other three modalities to get more accurate results of brain segmentation. We test this method on three datasets including IBSR dataset, MRBrainS13 dataset and MRBrainS18 datasets based on the deep voxelwise residual network VoxResNet\cite{Chen2016VoxResNet}.

\section{Method}
\subsection{Deformation Method}
The deformation method \cite{Liu1998An},\cite{Liao2000Level},\cite{Chen2015New} is derived from differential geometry. Consider $\mathrm{\Omega}$ and $\mathrm{\Omega}_t \subset \mathbb R^{2,3}$ with $0\leq{t}\leq{1}$, be moving (includes fixed) domains. Let $\pmb{v}(\pmb{x},t)$ be the velocity field on $\partial\mathrm{\Omega_t}$, where $\pmb{v}(\pmb{x},t)\cdot{\pmb{\mathrm{n}}}=0$ on any part of $\partial\mathrm{\Omega}_t$ with slippery-wall boundary conditions where $\pmb{\mathrm{n}}$ is the outward normal vector of $\partial\mathrm{\Omega}_t$. Given diffeomorphism $\pmb{\varphi}_0:\mathrm{\Omega}\rightarrow\mathrm{\Omega}_0$ and scalar function $f(\pmb{x},t)>0 \in C^1(\pmb{x},t)$ on the domain $\mathrm{\Omega}_t \times [0,1]$, such that

\begin{equation}\label{General1}
\begin{aligned}
&f(\pmb{x},0)=J(\pmb{\varphi}_0)\\
&\int_{\mathrm{\Omega}_t} \dfrac{1}{f(\pmb{x},t)}d\pmb{x} = |\mathrm{\Omega}_0|.
\end{aligned}
\end{equation}
A new (differ from $\pmb{\varphi}_0$) diffeomorphism $\pmb{\phi}(\pmb{\xi},t):\mathrm{\Omega}_0\rightarrow\mathrm{\Omega}_t$, such that  $J(\pmb{\phi}(\pmb{\xi},t)) =\text{det}\nabla(\pmb{\phi}(\pmb{\xi},t)) = f(\pmb{\phi}(\pmb{\xi},t),t)$, $\forall t \in [0,1]$,
can be constructed the following two steps:
\begin{itemize}
	\item First, determine $\pmb{u}(\pmb{x},t)$ on $\mathrm{\Omega}_t$ by solving
	\begin{equation}\label{General2}
	\left\{
	\begin{aligned}
	\text{div } \pmb{u}(\pmb{x},t)& = -\frac{\partial}{\partial t}(\dfrac{1}{f(\pmb{x},t)}) \\
	\text{curl } \pmb{u}(\pmb{x},t)& = 0\\
	\pmb{u}(\pmb{x},t)& = \dfrac{\pmb{v}(\pmb{x},t)}{f(\pmb{x},t)} \text{, on } \partial\mathrm{\Omega}_t
	\end{aligned}\right.
	\end{equation}
	\item Second, determine $\pmb{\phi}(\pmb{\xi},t)$ on $\mathrm{\Omega}_0$ by solving	
	\begin{equation}\label{General3}
	\left\{
	\begin{aligned}
	\frac{\partial \pmb{\phi}(\pmb{\xi},t)}{\partial t}& = f(\pmb{\phi}(\pmb{\xi},t),t) \pmb{u}(\pmb{\phi}(\pmb{\xi},t),t), \\
	\pmb{\phi}(\pmb{\xi},0)& = \pmb{\varphi}_{0}(\pmb{\xi})
	\end{aligned}\right.
	\end{equation}
\end{itemize}

For computational simplicity system (\ref{General2}) is modified into a Poisson equation as follows. Let $\pmb{u}(\pmb{x},t)= \nabla \pmb{w}(\pmb{x},t)$, then
\begin{equation}\label{General4}
\Delta \pmb{w}(\pmb{x},t) = \text{div } \nabla \pmb{w}(\pmb{x},t) = \text{div } \pmb{u}(\pmb{x},t) = -\frac{\partial}{\partial t}(\dfrac{1}{f(\pmb{x},t)})
\end{equation}
Hence, determination of $\pmb{u}(\pmb{x},t)$ is depended on $\pmb{w}(\pmb{x},t)$, which $\pmb{w}(\pmb{x},t)$ is cheaper to find.

%
%

\subsection{Experiments of Recovering Transformations}
Here we demonstrate some simulations of the deformation method. With both the Jacobian and curl term here in our algorithm, we can recover a transformation, which proves the algorithm is reliable and accurate in grid generation.
\subsubsection*{2D Example}

We design a recovering experiment to test the accuracy of our method and discover more details insides. We give the grid image of the map of (a)Guangdong, China, (b)Texas, USA and (c)a dog.
All images are given a nonlinear transformation $\mathbf{T_0}$ from the square($65\times65$)(phi1 and phi2).

%
\begin{figure}[H]
	\begin{center}
		\subfigure[]{\includegraphics[height=4.5cm,width=4.5cm]{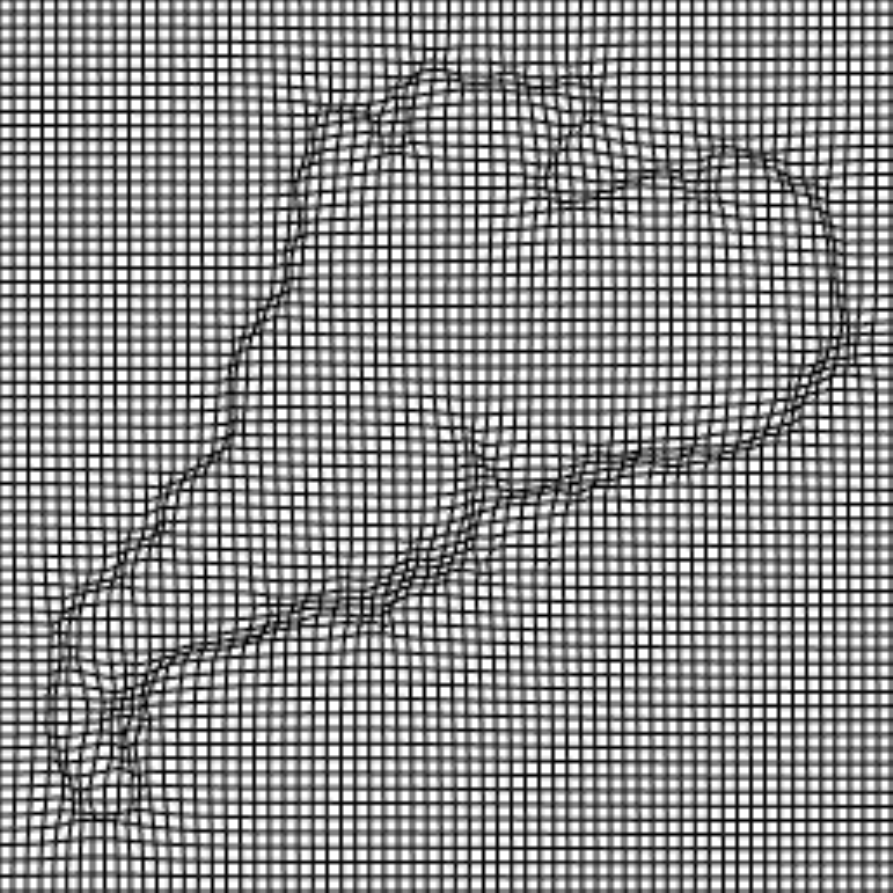}}
		\subfigure[]{\includegraphics[height=4.5cm,width=4.5cm]{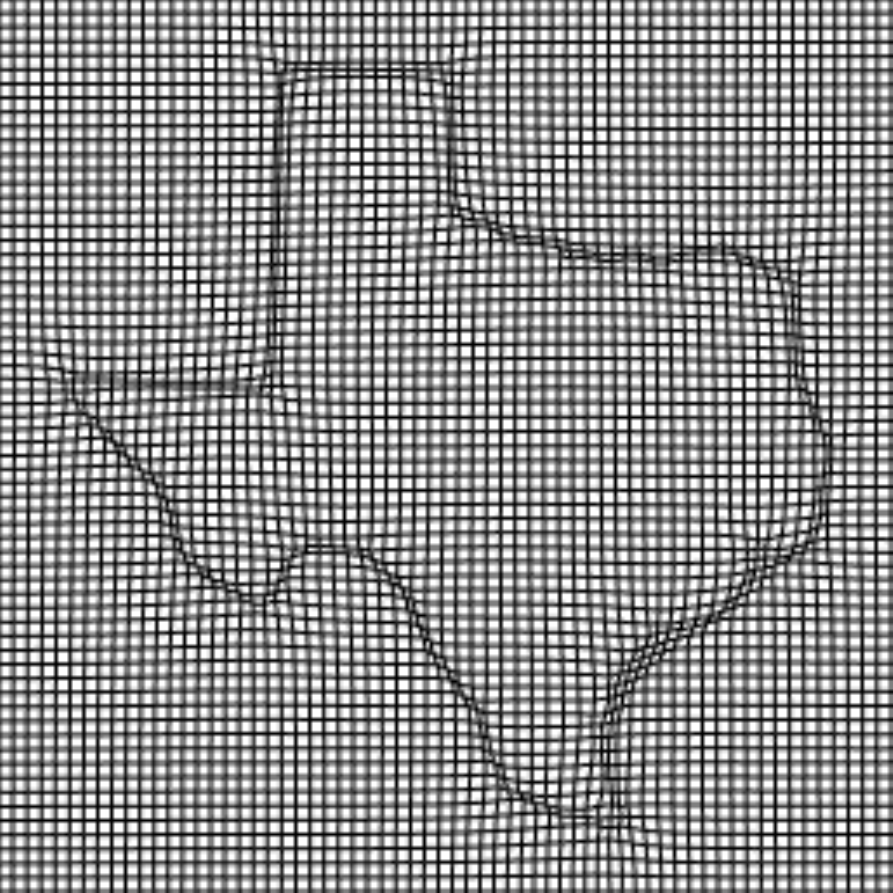}}
		\subfigure[]{\includegraphics[height=4.5cm,width=4.5cm]{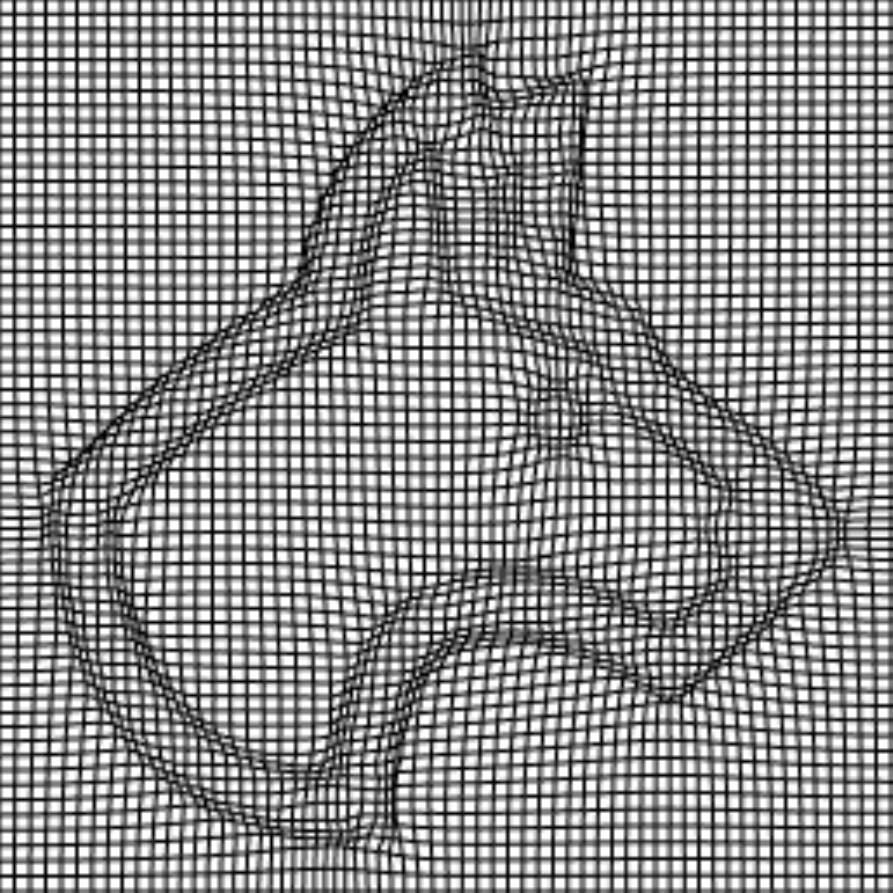}}
		\caption{Grids generated by setting intensity of images as $f(\pmb{x},0)$ of (\ref{General1}): (a) Grid: Guangdong, China,(b) Grid: Texas, USA (c) Grid: Dog year 2018.}
	\end{center}
\end{figure}

\subsubsection*{3D Example}

We also design a recovering experiment to brain MRI images and display them in three dimensions. Given a nonlinear transformation $\mathbf{T_0}$ from the cube $[1,176]\times[1,256]\times[1,256]$ (phi1, phi2 and phi3) to itself, we want to reconstruct $\mathbf{T_0}$ from its jacobian determinant and curl. We show the grid image from full scale view, half cut on x, y, z-axis view respectively.

\begin{figure}[H]
	\begin{center}
		\subfigure[]{\includegraphics[height=5.5cm,width=6.5cm]{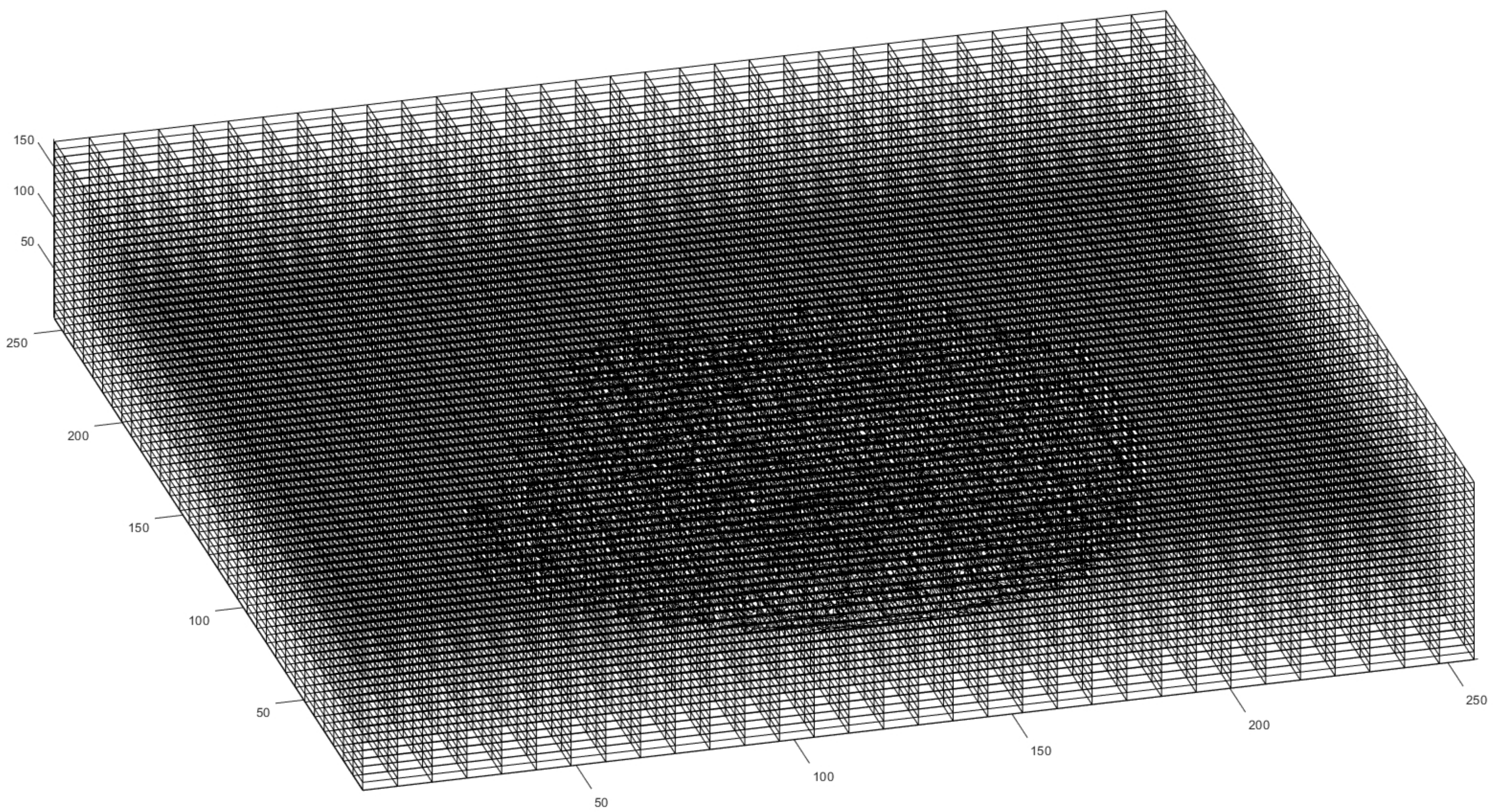}}
		\subfigure[]{\includegraphics[height=5.5cm,width=6.5cm]{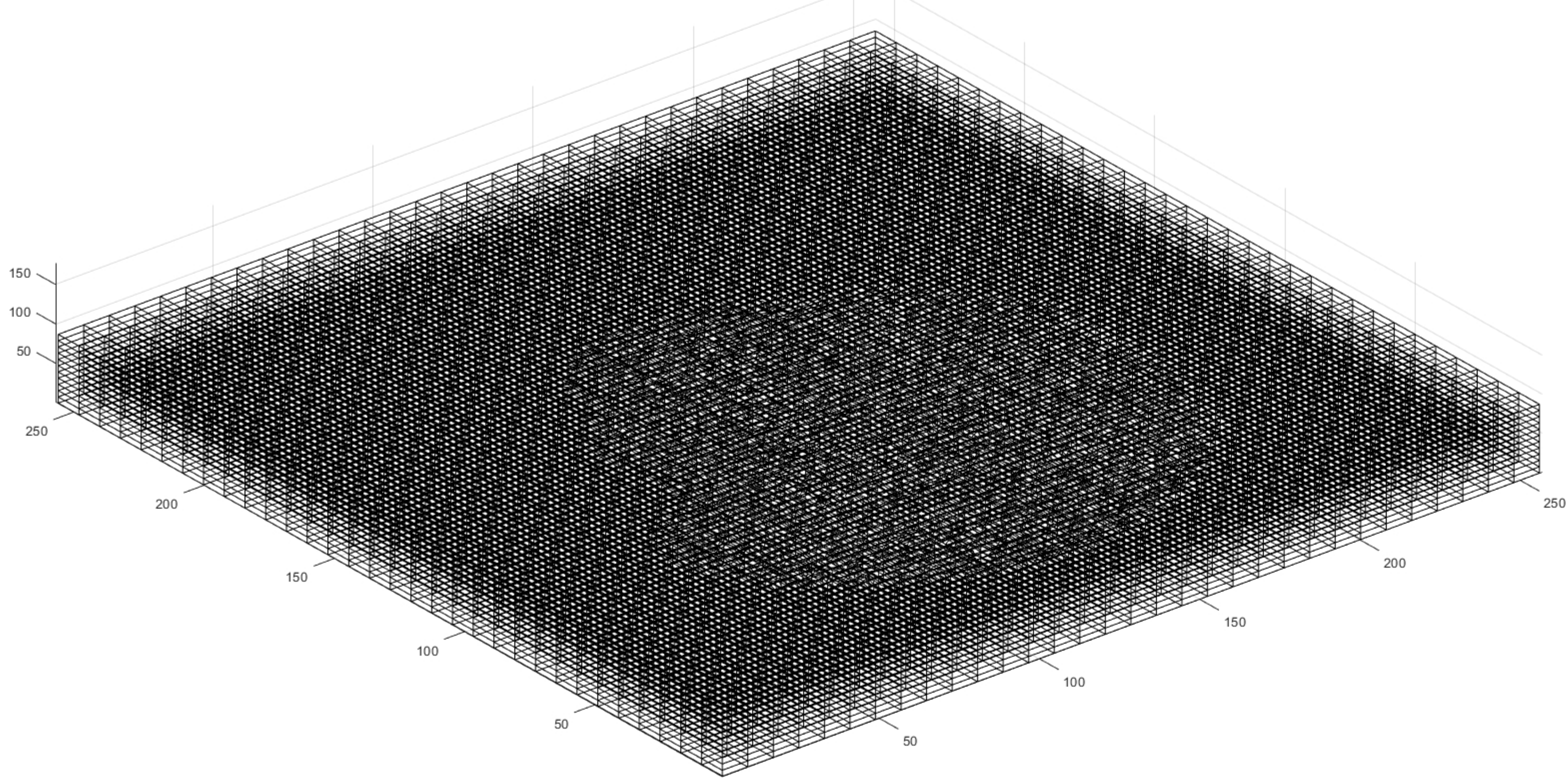}}
		\subfigure[]{\includegraphics[height=5.5cm,width=6.5cm]{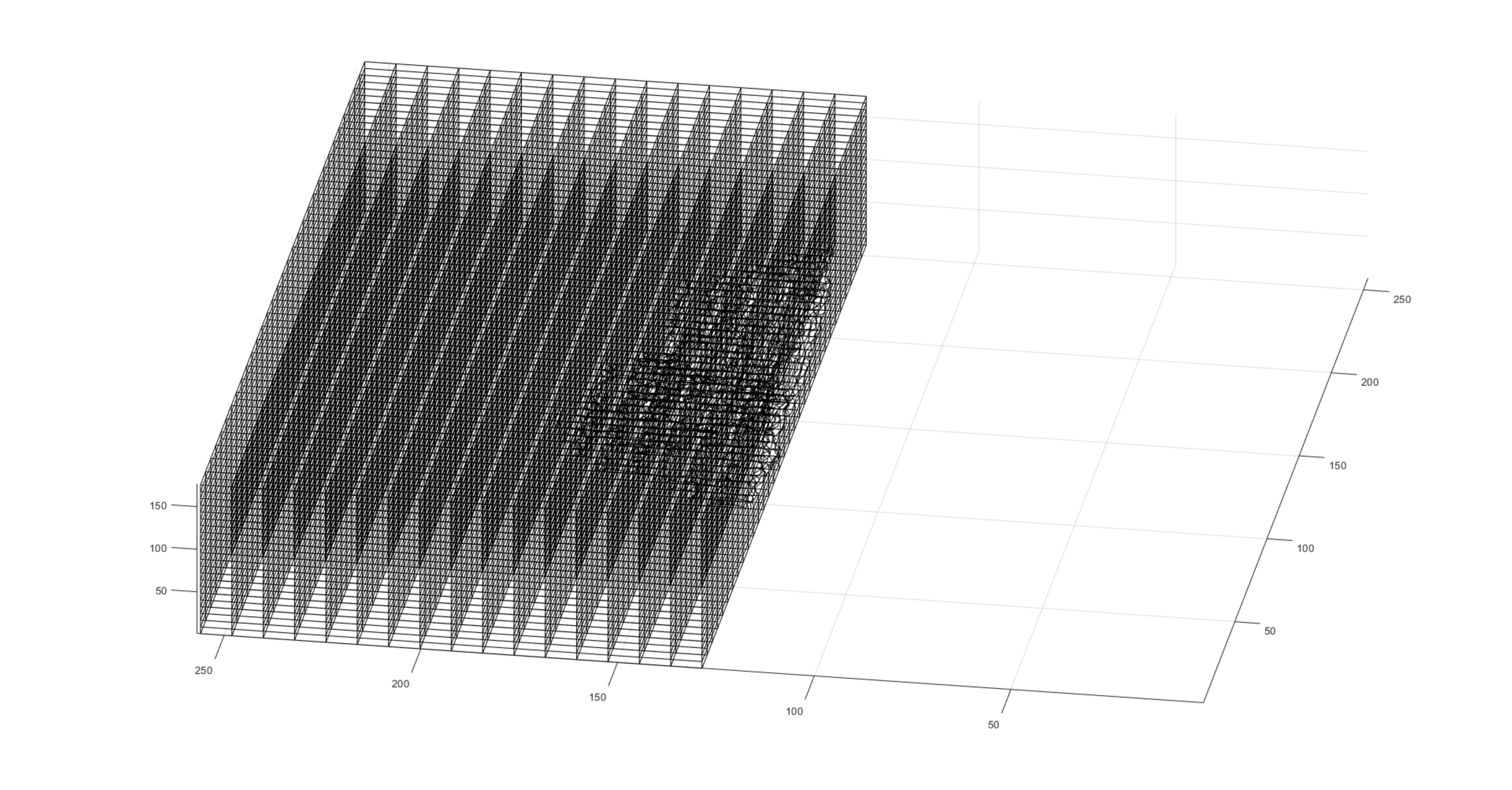}}
		\subfigure[]{\includegraphics[height=5.5cm,width=6.5cm]{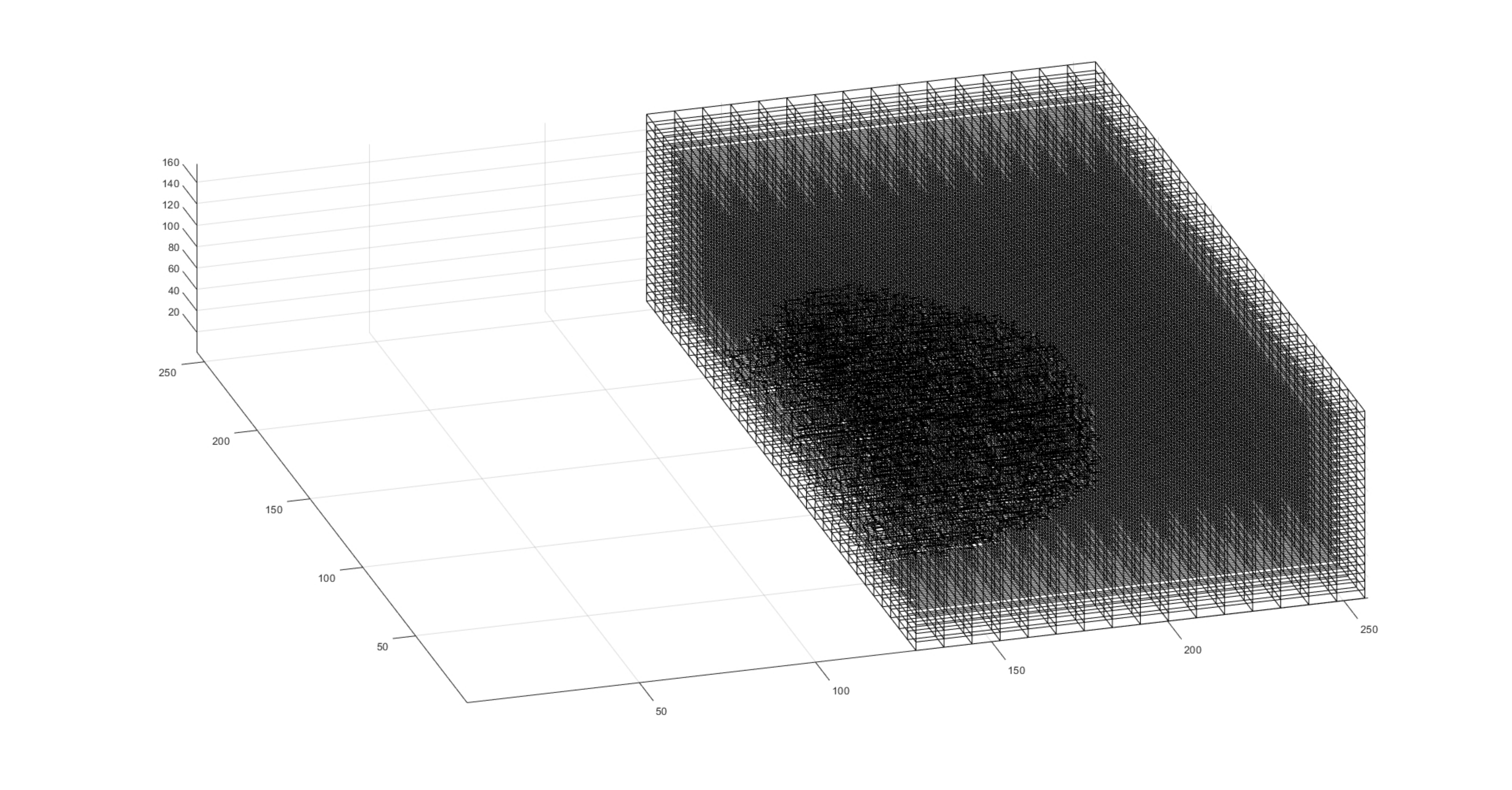}}
		\caption{Grid generated by a 3D brain image: (a) Full scale view, (b) Half cut on z-axis view, (c) Half cut on x-axis view,(d) Half cut on y-axis view.}
	\end{center}
\end{figure}

\subsection{Experiments of Grid Images Based on Brightness and Gradient}
In this experiment, we apply the method in 3.1 and 3.2 based on the brightness and its gradient of medical images. We generated corresponding grid images by calculating the brightness and gradient of brightness of brain MRI images, which further demonstrated that different brain MRI images could be differentiated by differential geometry. Figure 3(a) shows the original image(size 256$\times$256$\times$3).Figure 3(b)(c)(d) represent the grid images based on only gradient of brightness, only brightness(intensity) and both brightness and its gradient respectively.We can see that they reflect the morphological characteristics of CSF, GM, WM in MRI brain image. We used the differential geometric features of JD and CV as a method to measure the differences between different MRI images, and use this as CNN input to get more accurate results of brain segmentation.\\

\begin{figure}[H]
	\begin{center}
		\includegraphics[width=0.8\linewidth]{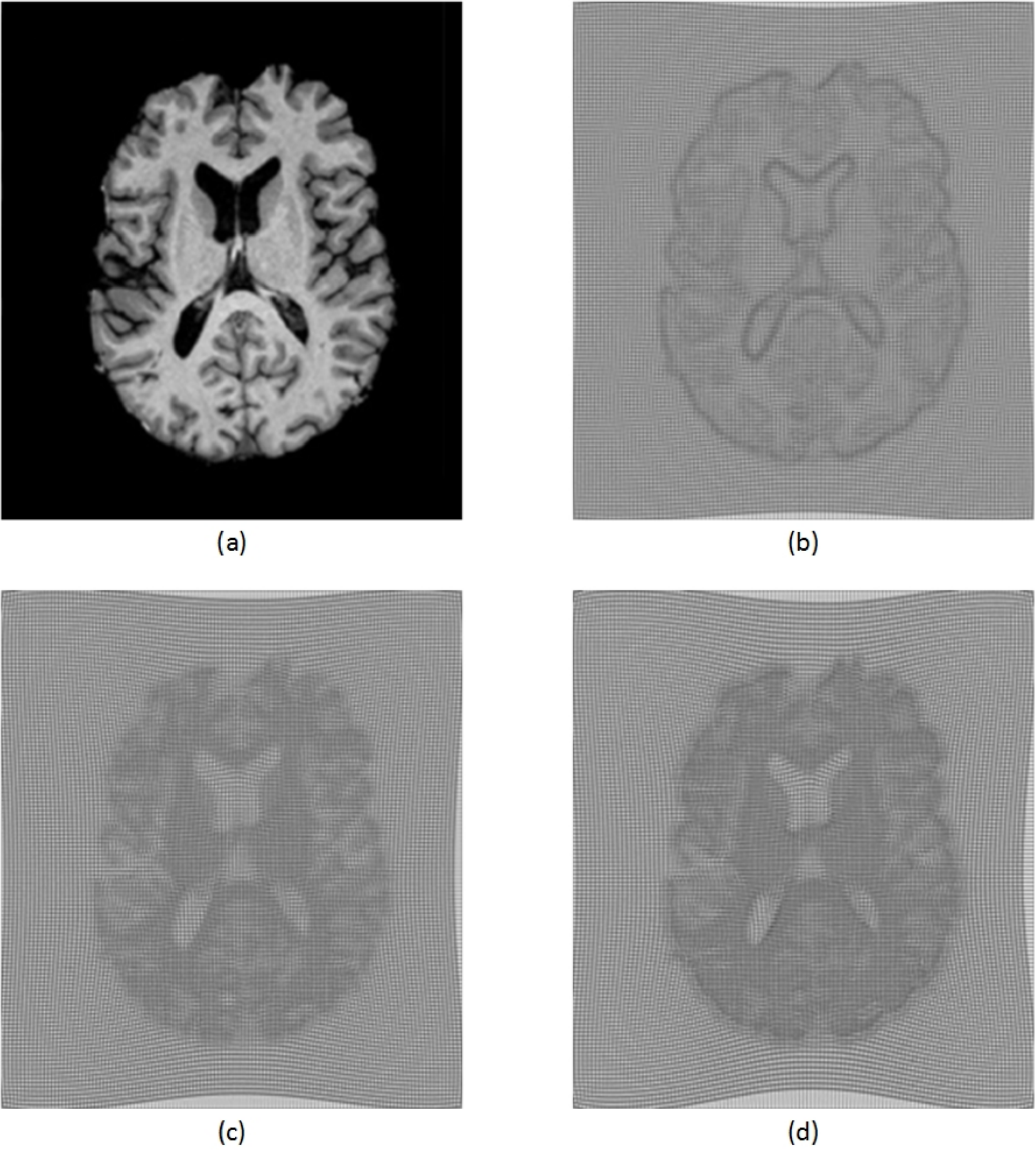}
		\caption{Two-dimensional grid image based on brain MRI image’s brightness and gradient of brightness: (a) The original image, (b) Grid image based on gradient of brightness, (c) Grid image based on brightness, (d) Grid images based on brightness and gradient of brightness.}
	\end{center}
\end{figure}

\subsection{Experiments of generated images based on JD and CV}
Based on the above finite difference and differential geometry theory, we use MATLAB to generate the grid images based on JD and CV of brain MRI images as 3.3 discribes.And we extract the images formed by JD and CV information from the grid images above, and saved it as nii image format. The following figures show in order are the original T1 image, the image formed by JD and the image formed by CV, which present geometric deformation features of the image, especially highlight the change of morphological features of CSF, GM, WM.\\

\begin{figure}[H]
	\begin{center}
		\includegraphics[width=1.\linewidth]{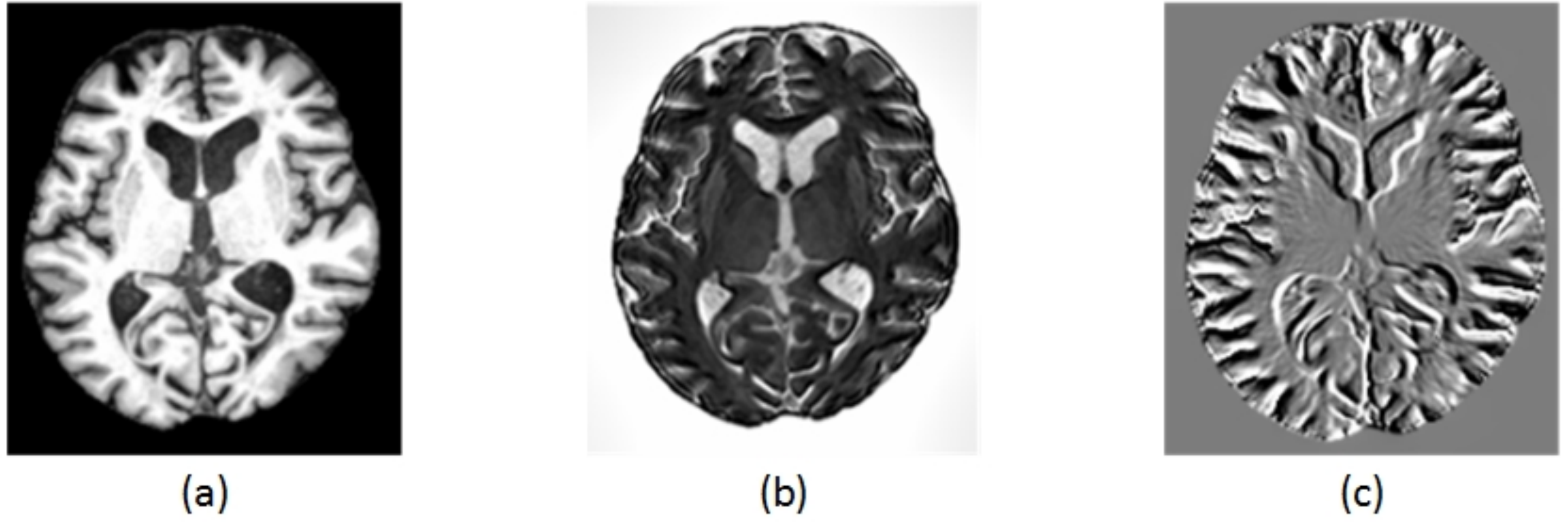}
		\caption{Two-dimensional generated images based on JD and CV: (a) The original T1 image, (b) The image formed by JD, (c) The image formed by CV.}
	\end{center}
\end{figure}

\subsection{Multi-modalities Information}
In medical image analysis, 3D volumetric data is usually obtained in a variety of ways for robust detection of different organizational structures. For example, three modalities including T1, T1-weighted inversion recovery (T1-IR) and T2-FLAIR are usually available in brain structure segmentation task \cite{Mendrik2015MRBrainS}. T1 image has good anatomical structure and T2 image can show good tissue lesions, T1-IR image has strong T1 contrast characteristics, and T2-FLAIR is often used to inhibit CSF issues. The main reason for obtaining multimodal images is that the information of multimodal data sets can supplement each other and provide robust diagnostic results. Thus, we concatenate these multi-modality data with JD and CV data as input, then in the process of network training, complementary information is combined and fused in an implicit way, which is more consistent than any single training method. The following figure shows each modality of brain image and the image formed by its JD and CV information.\\

\begin{figure}[H]
	\begin{center}
		\includegraphics[width=1.\linewidth]{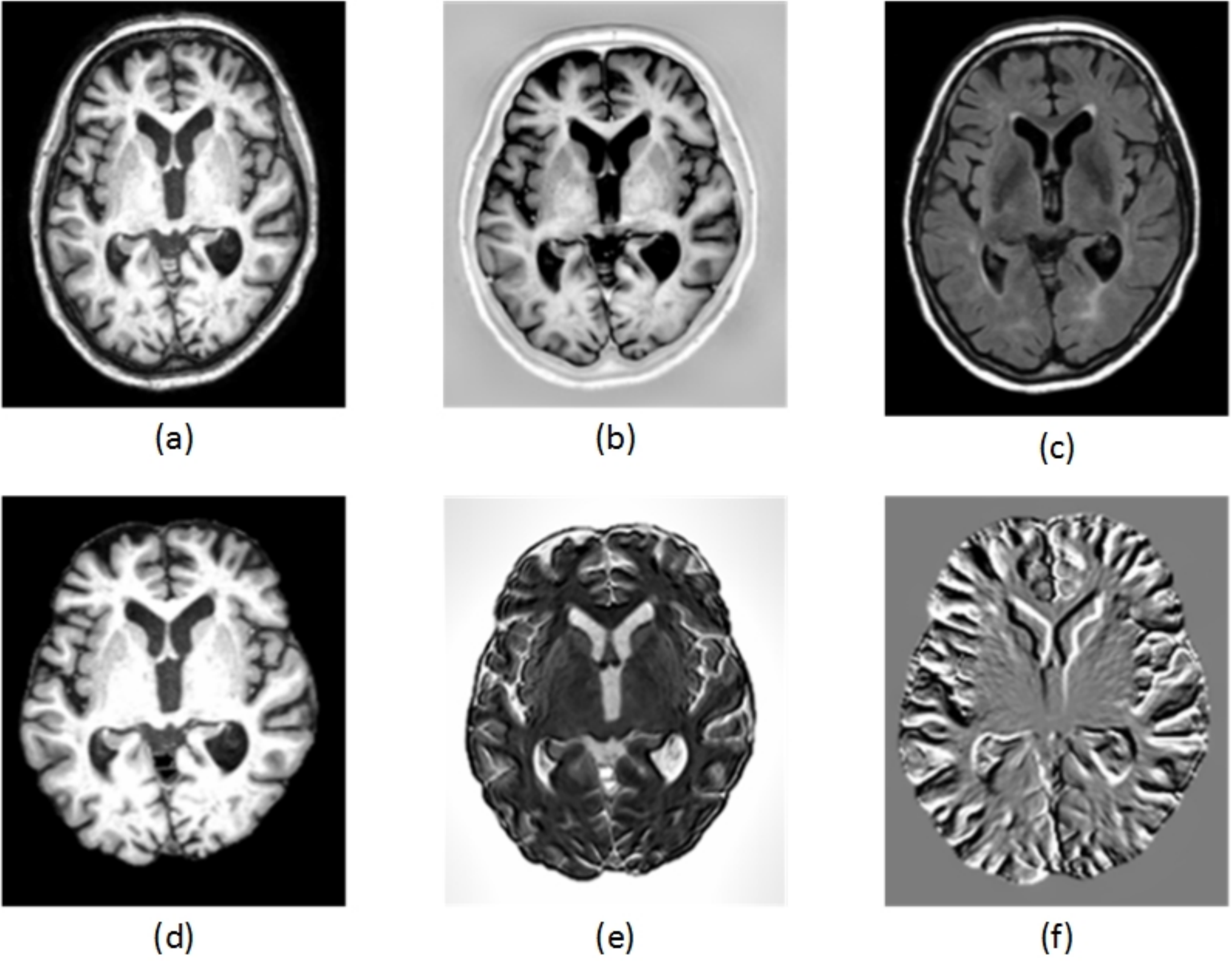}
		\caption{Each modality of brain image and the image formed by its JD and CV information: (a) T1 without skull stripping, (b) T1-IR without skull stripping, (c) T2-FLAIR without skull stripping, (d) T1 with skull stripping           (e) The image formed by JD of T1, (f) The image formed by CV of T1.}
	\end{center}
\end{figure}

\subsection{Proposed Framework}
Figure 6 shows the process framework of our proposed method. All modalities should be skull stripped and we can extract differential geometric features including the Jacobian determinant(JD) and the curl vector(CV) derived from T1 modality. After all modalities, including image labels(ground truth) have preprocessed, three modalities including T1-weighted, T1-IR and T2-FLAIR images with JD or CV image will be concatenated together as a new multi-modality. And it will be used as input of VoxResNet network. After training and testing, the predicted result with Ground Truth will be calculated for each tissue type(CSF, GM and WM), respectively, including DSC (Dice coefficient), HD ( Hausdorff distance) and AVD (Absolute Volume Difference) as evaluation criteria of segmentation. We will perform all the experiments based on this process.\\

\begin{figure}[H]
	\begin{center}
		\includegraphics[width=1.2\linewidth]{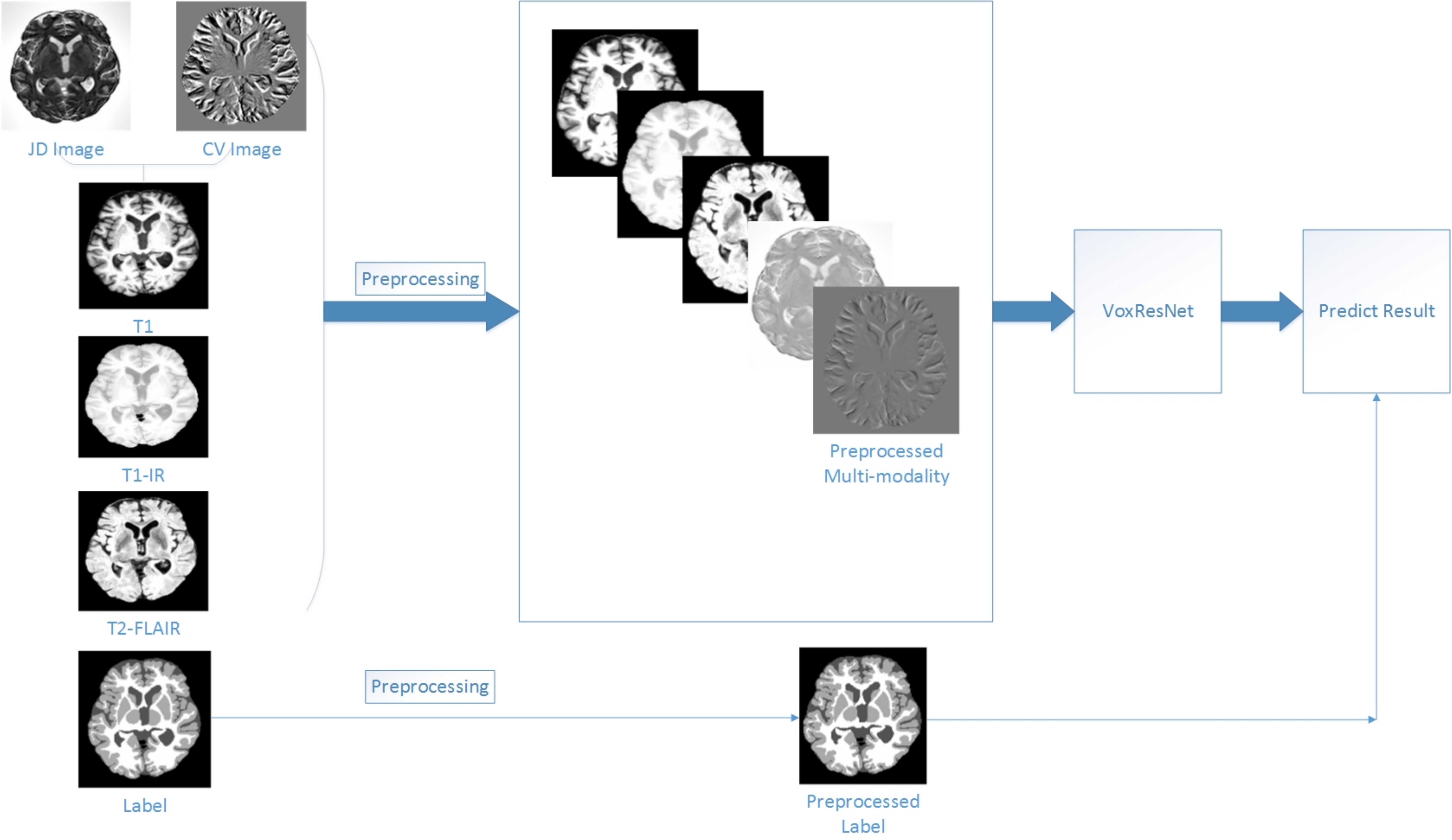}
		\caption{The process framework of our proposed method.}
	\end{center}
\end{figure}

\section{Results}

\subsection{Datasets and Pre-processing}

\subsubsection{Datasets}
\textbf{$\mathbf{IBSR}$} The Internet Brain Segmentation Repository (IBSR) provides manually-guided expert segmentation results along with magnetic resonance brain image data. Its purpose is to encourage the evaluation and development of segmentation methods. The dataset consists of 18 MRI volumes and the corresponding ground truth (GT) is provided. The data can be downloaded from \footnote{\url{https://www.nitrc.org/frs/?group_id=48}}.\\

\textbf{$\mathbf{MRBrainS}$} The aim of MRBrainS challenge is to segment brain into four-class structures, namely background, cerebrospinal fluid (CSF), gray matter (GM) and white matter (WM). Multi-sequence 3T MRI brain scans, including T1-weighted, T1-IR and T2-FLAIR are provided for each subject. Five brain MRI scans with manual segmentations are provided for training and 15 only MRI scans are provided for testing. The data can be downloaded from \footnote{\url{http://mrbrains18.isi.uu.nl/}} for MRBrainS18 dataset and \footnote{\url{http://mrbrains13.isi.uu.nl/}} for MRBrainS13 dataset.\\

\subsubsection{Image Preprocessing}	
Typical preprocessing steps for structural brain MRI include the following key steps \cite{Akkus2017Deep}: registration, skull stripping, bias field correction, intensity normalization and noise reduction. All the datasets we use are already skull stripped. In this paper, we subtract Gaussian smoothed image, and apply intensity normalization(z-scores) and Contrast-Limited Adaptive Histogram Equalization (CLAHE) for enhancing local contrast by \cite{Stollenga2015Parallel} in the pre-processing step. Then multiple input volumes pre-processed were used as input data in our experiments.\\

\subsection{Segmentation Evaluation Criteria}
In this paper, three metrics are used to evaluate the segmentation result \cite{Deng2018A} \cite{Sun2018A} : DSC (Dice coefficient), HD ( Hausdorff distance) and AVD (Absolute Volume Difference), which are calculated for each tissue type(CSF, GM and WM), respectively. First, DSC is the most widely used metric in the evaluation of medical volume segmentations. In addition to the direct comparison between automatic and ground truth segmentations, it is common to use the DSC to measure reproducibility (repeatability) \cite{Taha2015Metrics}. DSC is computed by:

\begin{equation}
DSC=\frac{2\times{TP}}{2\times{TP}+FP+FN}
\end{equation}

Where TP, FP and FN are the subjects of true positive, false positive and false negative predictions for the considered class.\\

Second, the distance between crisp volumes (HD) between two finite point sets A and B is defined by:

\begin{equation}
H(A,B)=max(h(A,B),h(B,A))
\end{equation}

Where $h(A,B)$ is called the directed HD and given by:
\begin{gather}
h(A,B)=max(a \in A)min(b \in B)\parallel a-b \parallel \\
h(B,A)=max(b \in B)min(a \in A)\parallel b-a \parallel
\end{gather}

Finally, the AVD is defined by:

\begin{equation}
AVD(A,B)=\frac{\parallel A-B \parallel}{\Sigma A}
\end{equation}

Where A is ground truth and B is predicted volume of one class.

\subsection{Experimental Results}

\subsubsection{Experiments on MRBrainS}
Regarding the evaluation of testing data, we compared our method with other modalities based on the network VoxResNet, including single modality(T1-weighted), three modalities(T1-weighted, T1-IR and T2-FLAIR), three modalities+JD modality, three modalities+CV modality and three modalities+JD+CV modality. In this experiment , we use subject 4,5,7,14,070 of MRBrainS18 as training set and subject 1,148 as testing set. We can see that combining the multi-modality information can improve the segmentation performance than that of single image modality. Three modalities gain a little improvement over single modality. While three modalities combined with JD, CV or both have excellent improvement over single or three modalities and three modalities with JD increases average Dice by about 1.5\%(0.8529 and 0.8677), especially gain much higher improvement by about 3.7\% increase (0.8490 and 0.8860) in WM tissue. This also can be seen from the improvement of HD and AVD(the smaller, the better) from the Table 2.\\

\begin{table}[H]
	\begin{center}
		\begin{tabular}{|*{10}{c|}}
			\toprule
			\multicolumn{1}{|c|}{Experiment} & \multicolumn{3}{|c|}{Dice} & \multicolumn{3}{|c|}{HD} & \multicolumn{3}{|c|}{AVD}\\
			\cmidrule{2-10}
			& CSF & GM & WM & CSF & GM & WM & CSF & GM & WM \\
			\midrule
			Single M(T1) & 0.8778 & 0.8138 & 0.8461 & 1 & 1.2071 & 1.9319 & \textbf{0.0373} & 0.0392 & 0.0751\\
			
			Three M & 0.8926 & 0.8172 & 0.8490 & 1 & 1.2071 & 1.8251 & 0.0458 & 0.0411 & 0.0937 \\
			
			Three M+JD & \textbf{0.8939} & 0.8232 & $\textbf{0.8860}$ & $\textbf{1}$ & $\textbf{1}$ & $\textbf{1.2071}$ & 0.0457 & $\textbf{0.0331}$ & 0.0484 \\
			
			Three M+CV & 0.8920 & 0.8217 & 0.8859 & 1 & 1 & 1.4142 & 0.0455 & 0.0373 & $\textbf{0.0457}$ \\
			
			Three M+JV & 0.8914 & $\textbf{0.8329}$ & 0.8701 & 1 & 1 & 1.5 & 0.0430 & 0.0508 & 0.0511 \\
			
			\bottomrule
		\end{tabular}
	\end{center}
	
	\caption{Test results of MRBrainS18 training set(4,5,7,14,070) for different experiments(DC:\%, HD: mm, AVD:\%).}
\end{table}

From the Table 2, we can see the subject 1(Dice:$\mathbf{0.8760}$, HD:$\mathbf{1.0829}$, AVD:$\mathbf{0.0294}$) of testing set obtain higher improvement than subject 148 in the performance in brain segmentation. And the experiment of three modalities+JD gains the best result with average Dice, HD, AVD by 0.8677,1.0690 and 0.0424.But three modalities combined with both JD and CV have a little decrease under three modalities combined with either JD or CV, maybe because of overfitting. \\

\begin{table}[H]
	\begin{center}
		\begin{tabular}{|*{10}{c|}}
			\toprule
			\multicolumn{1}{|c|}{Experiment} & \multicolumn{3}{|c|}{Dice} & \multicolumn{3}{|c|}{HD} & \multicolumn{3}{|c|}{AVD}\\
			\cmidrule{2-10}
			& 1 & 148 & average & 1 & 148 & average & 1 & 148 & average \\
			\midrule
			Single M(T1) & 0.8720 & 0.8198 & 0.8459 & 1.1381 & 1.6212 & 1.3797 & $\textbf{0.0156}$ & 0.0855 & 0.0505\\
			
			Three M & $\textbf{0.8796}$ & 0.8263 & 0.8529 & 1.1381 & 1.5501 & 1.3441 & 0.0273 & 0.0931 & 0.0602 \\
			
			Three M+JD & 0.8784 & 0.8570 & $\textbf{0.8677}$ & $\textbf{1}$ & $\textbf{1.1381}$ & $\textbf{1.0690}$ & 0.0333 & 0.0515 & $\textbf{0.0424}$ \\
			
			Three M+CV & 0.8742 & $\textbf{0.8589}$ & 0.8665 & 1.1381 & $\textbf{1.1381}$ & 1.1381 & 0.0366 & $\textbf{0.0491}$ & 0.0428 \\
			
			Three M+JV & 0.8756 & 0.8540 & 0.8648 & $\textbf{1}$ & 1.3333 & 1.1667 & 0.0341 & 0.0626 & 0.0483 \\
			
			average & $\textbf{0.8760}$ & 0.8432 & 0.8596 & $\textbf{1.0829}$ & 1.3562 & 1.2195 & $\textbf{0.0294}$ & 0.0684 & 0.0488 \\
			\bottomrule
		\end{tabular}
	\end{center}
	
	\caption{Testing set(1,148) results of MRBrainS18 training set(4,5,7,14,070) for different experiments(DC:\%, HD: mm, AVD:\%).}
\end{table}

Moreover, We use subject 1,3,4 of MRBrainS13 as training set and subject 2,5 as testing set and its results are shown in Table 3 and Table 4. We can get the same results as the MRBrainS18 dataset. The experiment of three modalities+CV gain the best Dice by 0.8272 and increase 1.7\% over only three modalities(Dice:0.8100).The subject 5 gain higher average Dice(0.8310) and HD(1.6697) but lower AVD(0.0607) than subject 2. We also compare the predicted results of subject 5 from the MRBrainS13 dataset and can be seen in Figure 8. We can get the same results as we discuss above. Three modalities combined with JD, CV or both have excellent improvement over only single or three modalities.\\
\begin{table}[H]
	\begin{center}
		\begin{tabular}{|*{10}{c|}}
			\toprule
			\multicolumn{1}{|c|}{Experiment} & \multicolumn{3}{|c|}{Dice} & \multicolumn{3}{|c|}{HD} & \multicolumn{3}{|c|}{AVD}\\
			\cmidrule{2-10}
			& CSF & GM & WM & CSF & GM & WM & CSF & GM & WM \\
			\midrule
			Single M(T1) & 0.8206 & 0.7855 & 0.7907 & 1.7071 & 1.5731 & 2.5 & 0.1072 & 0.1204 & 0.0751\\
			
			Three M & 0.8167 & 0.8115 & 0.8018 & $\textbf{1.4142}$ & $\textbf{1.4142}$ & $\textbf{2.4142}$ & 0.0574 & 0.0509 & 0.0679 \\
			
			Three M+JD & 0.8255 & 0.7984 & $\textbf{0.8078}$ & 1.7321 & 1.5731 & 2.5 & 0.1042 & 0.1067 & $\textbf{0.0084}$ \\
			
			Three M+CV & 0.8394 & $\textbf{0.8367}$ & 0.7985 & $\textbf{1.4142}$ & $\textbf{1.4142}$ & 2.5 & $\textbf{0.0369}$ & $\textbf{0.0098}$ & 0.0112 \\
			
			Three M+JV & $\textbf{0.8406}$ & 0.8343 & 0.8068 & $\textbf{1.4142}$ & $\textbf{1.4142}$ & 2.5811 & 0.0573 & 0.0310 & 0.0355 \\
			
			\bottomrule
		\end{tabular}
	\end{center}
	
	\caption{Test results of MRBrainS13 training set(1,3,4) for different experiments(DC:\%, HD: mm, AVD:\%).}
\end{table}

\begin{table}[H]
	\begin{center}
		\begin{tabular}{|*{10}{c|}}
			\toprule
			\multicolumn{1}{|c|}{Experiment} & \multicolumn{3}{|c|}{Dice} & \multicolumn{3}{|c|}{HD} & \multicolumn{3}{|c|}{AVD}\\
			\cmidrule{2-10}
			& 2 & 5 & average & 2 & 5 & average & 2 & 5 & average \\
			\midrule
			Single M(T1) & 0.7803 & 0.8176 & 0.7989 & 2.0488 & 1.8047 & 1.9267 & 0.0962 & 0.1057 & 0.1009\\
			
			Three M & 0.7975 & 0.8224 & 0.8100 & $\textbf{1.8856}$ & $\textbf{1.6095}$ & $\textbf{1.7475}$ & 0.0536 & 0.0639 & 0.0587 \\
			
			Three M+JD & 0.7919 & 0.8292 & 0.8106 & 2.1547 & 1.7154 & 1.9351 & 0.0715 & 0.0747 & 0.0731 \\
			
			Three M+CV & $\textbf{0.8078}$ & 0.8419 & 0.8249 & 1.9428 & 1.6095 & 1.7761 & $\textbf{0.0154}$ & $\textbf{0.0233}$ & $\textbf{0.0193}$ \\
			
			Three M+JV & 0.8072 & $\textbf{0.8437}$ & $\textbf{0.8272}$ & 1.9969 & 1.6095 & 1.8032 & 0.0465 & 0.0360 & 0.0413 \\
			
			average & 0.7969 & $\textbf{0.8310}$ & 0.8143 & 2.0058 & $\textbf{1.6697}$ & 1.8377 & $\textbf{0.0566}$ & 0.0607 & 0.0587 \\
			\bottomrule
		\end{tabular}
	\end{center}
	
	\caption{Testing set(2,5) results of MRBrainS13 training set(1,3,4) for different experiments(DC:\%, HD: mm, AVD:\%)}
\end{table}
From the red box in the Figure 7, We can see that (b)Single modality and (d)Three modalities+JD have a better result in WM and GM issue segmentation than other experiments. Other predicted results comparison of different experiments also can be seen in the Figure 7.
\begin{figure}[H]
	\begin{center}
		\includegraphics[width=1.\linewidth]{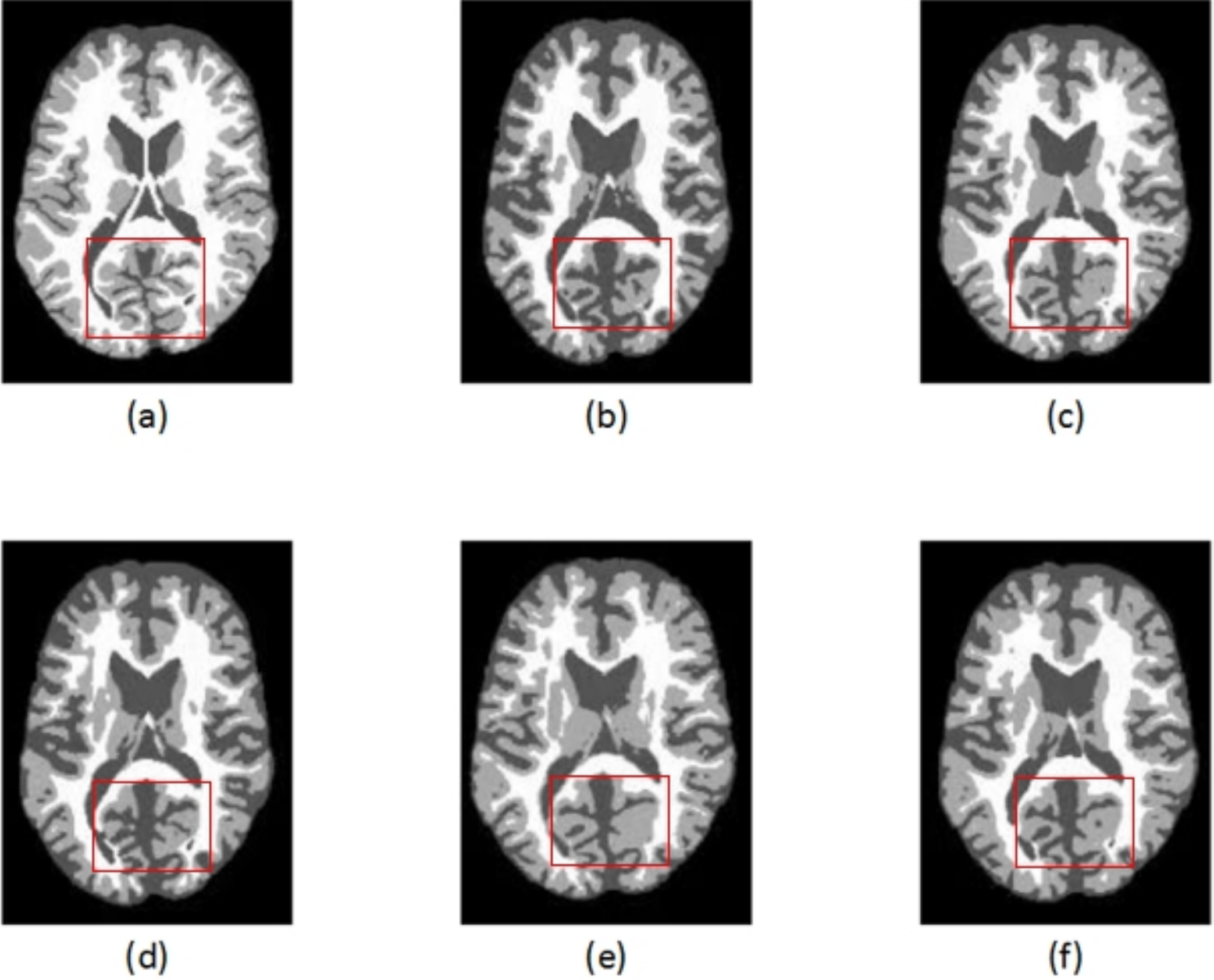}
		\caption{Predicted results comparison of different experiments: (a) Ground truth label, (b) Single modality(T1), (c) Three modalities, (d) Three modalities +JD, (e) Three modalities +CV, (f) Three modalities + JD + CV.}
	\end{center}
\end{figure}

\subsubsection{Experiments on IBSR}
In this experiment, we use subject 10,11,12,13,14 of IBSR as training set and subject 1-9 and 15-18 as testing set. we also compared our method with other modalities based on the network VoxResNet, including single modality(T1-weighted), single modality+JD modality, single modality+CV modality and single modalities+JD+CV modality. We can see that combining single modality with JD or CV information can improve the segmentation performance than that of single modality.\\

\begin{table}[H]
	\begin{center}
		\begin{tabular}{|*{10}{c|}}
			\toprule
			\multicolumn{1}{|c|}{Experiment} & \multicolumn{3}{|c|}{Dice} & \multicolumn{3}{|c|}{HD} & \multicolumn{3}{|c|}{AVD}\\
			\cmidrule{2-10}
			& CSF & GM & WM & CSF & GM & WM & CSF & GM & WM \\
			\midrule
			Single M(T1) & 0.7158 & 0.9137 & 0.9023 & 19.1106 & $\textbf{1.7693}$ & 1.5159 & 0.1877 & 0.0384 & 0.0461\\
			
			Single M+JD & $\textbf{0.7926}$ & $\textbf{0.9140}$ & 0.9008 & $\textbf{4.6863}$ & 1.7904 & 1.6379 & $\textbf{0.1584}$ & $\textbf{0.0296}$ & $\textbf{0.0450}$ \\
			
			Single M+CV & 0.7571 & 0.9113 & $\textbf{0.9024}$ & 6.8682 & 1.8951 & $\textbf{1.5101}$ & 0.1973 & 0.0327 & 0.0466 \\
			
			Single M+JV & 0.7612 & 0.9139 & 0.8888 & 5.0406 & 2.0204 & 1.8391 & 0.1879 & 0.0449 & 0.0652 \\
			
			\bottomrule
		\end{tabular}
	
	\end{center}
	
	\caption{Test results of IBSR training set(10-14) for different experiments(DC:\%, HD: mm, AVD:\%).}
\end{table}

Single modality combined with JD, CV or both have improvement over single modality and single modality with JD increases average Dice by about 2.5\%($\mathbf{0.8439}$ and $\mathbf{0.8691}$), especially gain much higher improvement by about 7.7\% increase ($\mathbf{0.7158}$ and $\mathbf{0.7926}$) in CSF tissue. This also can be seen from the improvement of HD($\mathbf{19.1106}$ and $\mathbf{4.6863}$)from the Table 5. But three modalities combined with both JD and CV also have a little decrease under three modalities combined with either JD or CV. From the Table 6, single modality with JD gets the best result with $\mathbf{0.8691}$,$\mathbf{2.7049}$,$\mathbf{0.0777}$ for average Dice, HD, AVD respectively.

\begin{table}[H]
	\begin{center}
		\begin{tabular}{|*{4}{c|}}
			\toprule
			\multicolumn{1}{|c|}{Experiment} & \multicolumn{1}{|c|}{Dice average} & \multicolumn{1}{|c|}{HD average} & \multicolumn{1}{|c|}{AVD average}\\
			
			\midrule
			Single modality(T1) & 0.8439 & 7.4653 & 0.0907 \\
			
			Single modality+JD & $\textbf{0.8691}$ & $\textbf{2.7049}$ & $\textbf{0.0777}$ \\
			
			Single modality+CV & 0.8569 & 3.4245 & 0.0922 \\
			
			Single modality+JV & 0.8546 & 2.9667 & 0.0993 \\
			
			\bottomrule
		\end{tabular}
	\end{center}
	
	\caption{Test average results of IBSR training set(10-14) for different experiments(DC:\%, HD: mm, AVD:\%).}
\end{table}

\subsubsection{Comparison with Other Experiments on MRBrainS18}
We also implement an additional experiment on the MRBrainS18 dataset. And we use subject 4,5,148,070 of MRBrainS18 as training set and subject 1,7,14 as testing set. Single modality with CV has the best results(Dice:$\mathbf{0.8710}$, HD:$\mathbf{1.0460}$, AVD:$\mathbf{0.0308}$), we can also see that combining single modality with JD or CV information can improve the segmentation performance than that of single modality. Single modality combined with JD, CV or both have a little improvement or not worse than three modalities(0.8619,0.8608,0.8710), which means one modality combined with its JD or CV information can replace the segmentation effect of three modalities, and patients only need to extract MRI images of T1 modality for diagnosis, which provides medical conveniences for doctor to diagnose.\\

\begin{table}[H]
	\begin{center}
		\begin{tabular}{|*{4}{c|}}
			\toprule
			\multicolumn{1}{|c|}{Experiment} & \multicolumn{1}{|c|}{Dice average} & \multicolumn{1}{|c|}{HD average} & \multicolumn{1}{|c|}{AVD average}\\
			
			\midrule
			Single modality(T1) & 0.8592 & 1.0920 & 0.0546\\
			
			Three modalities & 0.8619 & 1.1381 & 0.0630 \\
			
			Single modality+JD & 0.8608 & 1.0920 & 0.0517 \\
			
			Single modality+CV & $\textbf{0.8710}$ & $\textbf{1.0460}$ & $\textbf{0.0308}$ \\
			
			Single modality+JV &0.8634 & 1.1381 & 0.0333\\
			
			\bottomrule
		\end{tabular}
	\end{center}
	
	\caption{Test results of MRBrainS18 training set(4,5,148,070) for different experiments(DC:\%, HD: mm, AVD:\%).}
\end{table}

In Table 8, the subject 14 of testing set has better improvement results with average result (Dice:$\mathbf{0.8750}$, HD:$\mathbf{1.0552}$, AVD:$\mathbf{0.0418}$)than subject 1 and 7.

\begin{table}[H]
	\begin{center}
		\begin{tabular}{|*{10}{c|}}
			\toprule
			\multicolumn{1}{|c|}{Experiment} & \multicolumn{3}{|c|}{Dice} & \multicolumn{3}{|c|}{HD} & \multicolumn{3}{|c|}{AVD}\\
			\cmidrule{2-10}
			& 1 & 7 & 14 & 1 & 7 & 14 & 1 & 7 & 14 \\
			\midrule
			Single M(T1) & 0.8543 & 0.8445 & $\textbf{0.8788}$ & 1.1381 & 1.1381 & $\textbf{1}$ & 0.0494 & 0.0711 & 0.0432\\
			
			Three M & 0.8514 & 0.8584 & 0.8759 & 1.1381 & 1.1381 & 1.1381 & 0.0890 & 0.0558 & 0.0442\\
			
			Single M+JD & 0.8682 & 0.8467 & 0.8674 & 1.1381 & 1.1381 & $\textbf{1}$ & 0.0295 & 0.0625 & 0.0631\\
			
			Single M+CV & $\textbf{0.8736}$ & $\textbf{0.8627}$ & 0.8766 & $\textbf{1}$ & $\textbf{1.1381}$ & $\textbf{1}$ & $\textbf{0.0191}$ & 0.0367 & 0.0366\\
			
			Single M+JV & 0.8618 & 0.8522 & 0.8762 & 1.1381 & 1.1381 & 1.1381 & 0.0442 & $\textbf{0.0337}$ & $\textbf{0.0220}$\\
			average & 0.8619 & 0.8529 & $\textbf{0.8750}$ & 1.1105 & 1.1381 & $\textbf{1.0552}$ & 0.0462 & 0.0520 & $\textbf{0.0418}$\\
			\bottomrule
		\end{tabular}
	\end{center}
	
	\caption{Testing set(1,7,14) results of MRBrainS18 training set(4,5,148,070) for different experiments(DC:\%, HD: mm, AVD:\%).}
\end{table}

\subsubsection{Comparison between VoxResNet and U-Net on MRBrainS}

In the experiment, we compare the segmentation performance of our method in two networks, VoxResNet and U-Net network. We also use subject 4,5,7,14,070 of MRBrainS18 as training set and subject 1,148 as testing set. The results show VoxResNet has a little better performance with average results(Dice:$\mathbf{0.8596}$, HD:$\mathbf{1.2195}$, AVD:$\mathbf{0.0488}$)than U-Net network with our method in brain MRI segmentation and increase average Dice by about 1\% and this result can be also seen from HD and AVD from Table 9, especially in the case of three modalities with JD, CV or both. While VoxResNet has weaker performance in the case of single or three modalities. \\

\begin{table}[H]
	\begin{center}
		\begin{tabular}{|*{7}{c|}}
			\toprule
			\multicolumn{1}{|c|}{Experiment} & \multicolumn{2}{|c|}{Dice} & \multicolumn{2}{|c|}{HD} & \multicolumn{2}{|c|}{AVD}\\
			\cmidrule{2-7}
			& VoxResNet & U-Net & VoxResNet & U-Net & VoxResNet & U-Net \\
			\midrule
			Single modality(T1) & 0.8459 & $\textbf{0.8669}$ & $\textbf{1.3797}$ & 1.9917 & 0.0505 & $\textbf{0.0397}$\\
			
			Three modalities & 0.8529 & $\textbf{0.8548}$ & $\textbf{1.3441}$ & 2.2270 & 0.0602 & $\textbf{0.0479}$\\
			
			Three modalities+JD & $\textbf{0.8677}$ & 0.8526 & $\textbf{1.0690}$ & 2.1578 & $\textbf{0.0424}$ & 0.0464 \\
			
			Three modalities+CV & $\textbf{0.8665}$ & 0.8547 & $\textbf{1.1381}$ & 2.1075 & $\textbf{0.0428}$ & 0.0506\\
			
			Three modalities+JV & $\textbf{0.8648}$ & 0.8566 & $\textbf{1.1667}$ & 2.0191 & 0.0483 & $\textbf{0.0468}$ \\
			average & $\textbf{0.8596}$ & 0.8571 & $\textbf{1.2195}$ & 2.1006 & 0.0488 & $\textbf{0.0463}$ \\
			
			\bottomrule
		\end{tabular}
	\end{center}
	
	\caption{Test results comparison of MRBrainS18 training set(4,5,7,14,070) for different experiments between VoxResNet and U-Net(DC:\%, HD: mm, AVD:\%).}
\end{table}

We also get the same results based on MRBrainS13 by using subject 1,3,4 of MRBrainS13 as training set and subject 2,5 as testing set from Table 9.But VoxResNet has much higher performance with average results(Dice:$\mathbf{0.8143}$, HD:$\mathbf{1.8377}$, AVD:$\mathbf{0.0587}$)than U-Net network(Dice:$\mathbf{0.5635}$, HD:$\mathbf{4.7994}$, AVD:$\mathbf{0.2385}$) with our method in MRBrainS13.

\begin{table}[H]
	\begin{center}
		\begin{tabular}{|*{7}{c|}}
			\toprule
			\multicolumn{1}{|c|}{Experiment} & \multicolumn{2}{|c|}{Dice} & \multicolumn{2}{|c|}{HD} & \multicolumn{2}{|c|}{AVD}\\
			\cmidrule{2-7}
			& VoxResNet & U-Net & VoxResNet & U-Net & VoxResNet & U-Net \\
			\midrule
			Single modality(T1) & $\textbf{0.7989}$ & 0.6469 & $\textbf{1.9267}$ & 3.9520 & $\textbf{0.1009}$ & 0.1297\\
			
			Three modalities & $\textbf{0.8100}$ & 0.5448 & $\textbf{1.7475}$ & 4.8834 & $\textbf{0.0587}$ & 0.2530\\
			
			Three modalities+JD & $\textbf{0.8106}$ & 0.5438 & $\textbf{1.9351}$ & 4.9356 & $\textbf{0.0731}$ & 0.2575 \\
			
			Three modalities+CV & $\textbf{0.8249}$ & 0.5220 & $\textbf{1.7761}$ & 5.2546 & $\textbf{0.0193}$ & 0.2899\\
			
			Three modalities+JV & $\textbf{0.8272}$ & 0.5601 & $\textbf{1.8032}$ & 4.9716 & $\textbf{0.0413}$ & 0.2626 \\
			average & $\textbf{0.8143}$ & 0.5635 & $\textbf{1.8377}$ & 4.7994 & $\textbf{0.0587}$ & 0.2385 \\
			\bottomrule
		\end{tabular}
	\end{center}
	
	\caption{Test average results comparison of MRBrainS13 training set(1,3,4) for different experiments between VoxResNet and U-Net(DC:\%, HD: mm, AVD:\%).}
\end{table}

\section{Implementation Details}
Our method was implemented using MATLAB and a flexible framework neural networks named Chainer in Python. we used MATLAB to generate the images formed by JD and CV information based on brain MRI images, and saved it as nii image format. It took about 8 hours to train the network while less than 3 minutes for processing each test volume(size $240\times240\times48$) using one NVIDIA Quadro P2000 GPU. Due to the limited GPU memory, we cropped volumetric regions(size $80\times80\times80\times{m}$, m is the number of image modalities and set as 2,6,8,8,10 for single modality, three modalities, three modalities+JD, three modalities+CV, three modalities+JD+CV respectively in our experiments) for the input into the network. This was implemented in an on-the-fly way during the training and the probability map of whole volume was generated in an overlap-tiling strategy for stitching the sub-volume results.\\

\section{Conclusions}
In this paper, we use the differential geometric information including JD and CV derived from T1 modality as MRI image features, and use them as one CNN channel with other three modalities (T1-weighted, T1-IR and T2-FLAIR) to get more accurate results of brain segmentation. We test this method on two datasets including IBSR dataset and MRBrainS datasets based on VoxResNet network, and obtain excellent improvement on the two datasets. Moreover, we discuss that one modality combined with its JD or CV information can replace the segmentation effect of three modalities, which can provide medical conveniences for doctor to diagnose. Finally, we compare the segmentation performance of our method in two networks, VoxResNet and U-Net network. We believe the proposed method can advance the performance in brain segmentation and clinical diagnosis. In the future, we will investigate the performance of our method on more object detection and segmentation tasks from 3D volumetric data.\\

\section{Acknowledgments}
The authors would like to thank Yang Deng and Yao Sun of Graduate School at Shenzhen, Tsinghua University, Mingwang Zhu of Beijing Sanbo Brain Hospital for their technical support.\\

\section{Appendix: Other Experimental Results on MRBrainS}
We change the training set and testing set of MRBrainS18 dataset and the results in Table 11 and 12 show our method has the same improvement as previous experiments in 4.3.1. The best results of different experiments are colored by black. The experiment of three modalities with JD has the best result with average Dice, HD, AVD by 0.8779, 1, 0.0297 respectively(not shown in Table 11).

\begin{table}[H]
	\begin{center}
		\begin{tabular}{|*{10}{c|}}
			\toprule
			\multicolumn{1}{|c|}{Experiment} & \multicolumn{3}{|c|}{Dice} & \multicolumn{3}{|c|}{HD} & \multicolumn{3}{|c|}{AVD}\\
			\cmidrule{2-10}
			& CSF & GM & WM & CSF & GM & WM & CSF & GM & WM\\
			\midrule
			Single M(T1) & 0.8745 & 0.8298 & 0.8733 & 1 & 1 & 1.2761 & 0.0459 & 0.0330 & 0.0849 \\
			
			Three M & 0.8796 & 0.8436 & 0.8624 & 1 & 1 & 1.4142 & 0.0580 & 0.0766 & 0.0543 \\
			
			Three M+JD & $\textbf{0.8934}$ & $\textbf{0.8531}$ & $\textbf{0.8872}$ & $\textbf{1}$ & $\textbf{1}$ & $\textbf{1}$ & 0.0413 & $\textbf{0.0224}$ & $\textbf{0.0255}$ \\
			
			Three M+CV & 0.8927 & 0.8371 & 0.8764 & 1 & 1 & 1.2761 & $\textbf{0.0408}$ & 0.0362 & 0.0684 \\
			
			Three M+JV & 0.8911 & 0.8481 & 0.8854 & 1 & 1 & 1.1381 & 0.0442 & 0.0245 & 0.0476 \\
			
			\bottomrule
		\end{tabular}
	\end{center}
	\caption{Test results of MRBrainS18 training set(4,5,148,070) for different experiments(DC:\%, HD: mm, AVD:\%).}
\end{table}

\begin{table}[H]
	\begin{center}
		\begin{tabular}{|*{10}{c|}}
			\toprule
			\multicolumn{1}{|c|}{Experiment} & \multicolumn{3}{|c|}{Dice} & \multicolumn{3}{|c|}{HD} & \multicolumn{3}{|c|}{AVD}\\
			\cmidrule{2-10}
			& 1 & 7 & 14 & 1 & 7 & 14 & 1 & 7 & 14 \\
			\midrule
			Single M(T1) & 0.8543 & 0.8445 & 0.8788 & 1.1381 & 1.1381 & $\textbf{1}$ & 0.0494 & 0.0711 & 0.0432\\
			
			Three M & 0.8514 & 0.8584 & 0.8759 & 1.1381 & 1.1381 & 1.1381 & 0.0890 & 0.0558 & 0.0442\\
			
			Three M+JD & $\textbf{0.8775}$ & $\textbf{0.8724}$ & 0.8837 & $\textbf{1}$ & $\textbf{1}$ & $\textbf{1}$ & $\textbf{0.0279}$ & $\textbf{0.0291}$ & 0.0323\\
			
			Three M+CV & 0.8664 & 0.8573 & 0.8824 & 1.1381 & 1.1381 & $\textbf{1}$ & 0.0373 & 0.0586 & 0.0495\\
			
			Three M+JV & 0.8721 & 0.8623 & $\textbf{0.8902}$ & $\textbf{1}$ & 1.1381 & $\textbf{1}$ & 0.0416 & 0.0453 & $\textbf{0.0294}$\\
			
			average & 0.8643 & 0.8590 & $\textbf{0.8822}$ & 1.0829 & 1.1105 & $\textbf{1.0276}$ & 0.0490 & 0.0520 & $\textbf{0.0397}$\\
			\bottomrule
		\end{tabular}
	\end{center}
	\caption{Testing set(1,7,14) results of MRBrainS18 training set(4,5,148,070) for different experiments(DC:\%, HD: mm, AVD:\%).}
\end{table}

Table 13 and Table 14 show Single modality combined with JD, CV or both have a little improvement or not worse than three modalities of MRBrainS13 dataset as 4.3.3 describes. And the subject 5 of testing set has better improvement results than subject 1 with average Dice, HD, AVD by 0.8493, 1.5640, 0.0385 respectively from Table 14.

\begin{table}[H]
	\begin{center}
		\begin{tabular}{|*{4}{c|}}
			\toprule
			\multicolumn{1}{|c|}{Experiment} & \multicolumn{1}{|c|}{Dice average} & \multicolumn{1}{|c|}{HD average} & \multicolumn{1}{|c|}{AVD average}\\
			\midrule
			Single modality(T1) & 0.7840 & 2.1979 & 0.0571\\
			
			Three modalities & 0.7856 & 2.0426 & 0.0590 \\
			
			Single modality+JD & $\textbf{0.7889}$ & 2.0242 & 0.0713 \\
			
			Single modality+CV & 0.7870 & $\textbf{1.9538}$ & $\textbf{0.0512}$ \\
			
			Single modality+JV & 0.7805 & 2.1343 & 0.0734 \\
			
			\bottomrule
		\end{tabular}
	\end{center}
	\caption{Test average results of MRBrainS13 training set(2,3,4) for different experiments(DC:\%, HD: mm, AVD:\%).}
\end{table}

\begin{table}[H]
	\begin{center}
		\begin{tabular}{|*{7}{c|}}
			\toprule
			\multicolumn{1}{|c|}{Experiment} & \multicolumn{2}{|c|}{Dice} & \multicolumn{2}{|c|}{HD} & \multicolumn{2}{|c|}{AVD}\\
			\cmidrule{2-7}
			& 1 & 5 & 1 & 5 & 1 & 5 \\
			\midrule
			Single modality(T1) & 0.7148 & $\textbf{0.8531}$ & 2.7864 & 1.6095 & 0.0748 & 0.0393 \\
			
			Three modalities & 0.7183 & 0.8529 & 2.6139 & $\textbf{1.4714}$ & 0.1056 & $\textbf{0.0123}$ \\
			
			Single modality+JD & 0.7258 & 0.8521 & 2.4389 & 1.6095 & 0.1014 & 0.0412 \\
			
			Single modality+CV & $\textbf{0.7282}$ & 0.8459 & $\textbf{2.3874}$ & 1.5202 & $\textbf{0.0492}$ & 0.0533 \\
			
			Single modality+JV & 0.7184 & 0.8425 & 2.6593 & 1.6095 & 0.1005 & 0.0463 \\
			average & 0.7211 & $\textbf{0.8493}$ & 2.5772 & $\textbf{1.5640}$ & 0.0863 & $\textbf{0.0385}$ \\
			\bottomrule
		\end{tabular}
	\end{center}
	\caption{Testing set(1,5) results of MRBrainS13 training set(2,3,4) for different experiments(DC:\%, HD: mm, AVD:\%).}
\end{table}

\bibliographystyle{IEEEtranS}
\bibliography{D3D_DRL}

\end{document}